\documentclass{article}
\usepackage{spconf,amsmath,graphicx}
\usepackage{algorithm}
\usepackage[noend]{algpseudocode}
\usepackage{makecell}
\usepackage{booktabs}
\usepackage{comment}
\usepackage{caption}
\title{Towards Unsupervised Speech-to-Text Translation}

\name{Yu-An Chung\quad Wei-Hung Weng\quad Schrasing Tong\quad James Glass}
\address{Computer Science and Artificial Intelligence Laboratory\\
    Massachusetts Institute of Technology\\
    Cambridge, MA 02139, USA\\
    \tt{\{andyyuan,ckbjimmy,st9,glass\}@mit.edu}}

\begin{document}
%\ninept

\maketitle

\begin{abstract}
We present a framework for building speech-to-text translation~(ST) systems using only monolingual speech and text corpora, in other words, speech utterances from a source language and independent text from a target language.
As opposed to traditional cascaded systems and  end-to-end architectures, our system does not require any labeled data~(i.e., transcribed source audio or parallel source and target text corpora) during training, making it especially applicable to language pairs with very few or even zero bilingual resources.
% To overcome the lack of parallel corpora,
The framework initializes the ST system with a \textit{cross-modal} bilingual dictionary inferred from the monolingual corpora, that maps every source speech segment corresponding to a spoken word to its target text translation.
For unseen source speech utterances, the system first performs word-by-word translation on each speech segment in the utterance.
The translation is improved by leveraging a language model and a sequence denoising autoencoder to provide prior knowledge about the target language.
Experimental results show that our unsupervised system achieves comparable BLEU scores to supervised end-to-end models despite the lack of supervision.
We also provide an ablation analysis to examine the utility of each component in our system.
\end{abstract}

\begin{keywords}
speech-to-text translation, unsupervised speech processing, speech2vec, bilingual lexicon induction
\end{keywords}

\section{Introduction}
Conventional speech-to-text translation~(ST) systems, which typically cascade automatic speech recognition~(ASR) and machine translation~(MT), impose significant requirements on training data.
%~\cite{waibel2008spoken}.
They usually require hundreds of hours of transcribed audio and millions of words of parallel text from the source and target languages to train individual components,
which makes it difficult to use this approach on low-resource languages.
Although recent works have shown the feasibility of building end-to-end systems that directly translate source speech to target text without using any intermediate source language transcriptions, they still require data in the form of source audio paired with target text translations for end-to-end training~\cite{weiss2017sequence,bansal2018low,berard2018end,berard2016listen}.

In contrast to ST, which requires paired data for training, recent research in MT has explored fully unsupervised settings---relying only on monolingual corpora from each language.
They have shown that unsupervised MT models can achieve comparable~(sometimes even superior) results to supervised ones~\cite{lample2018phrase,artetxe2018unsupervised}.
%~\cite{lample2018unsupervised,artetxe2018unsupervised2}
A key principle behind these unsupervised MT approaches is to initialize a MT model with a bilingual dictionary inferred from monolingual corpora, without using cross-lingual signals~\cite{conneau2018word,artetxe2018robust}.
Given a source word, the initial MT model is able to perform word-by-word translation by looking up the dictionary, and can be further improved by leveraging other techniques such as back translation~\cite{sennrich2016improving}.

Recently,~\cite{chung2018unsupervised} showed that these unsupervised bilingual dictionary induction algorithms could also be applied to scenarios where the source and target corpora are of different modalities, namely speech and text.
The learned \textit{cross-modal} bilingual dictionary, as we will show in this paper, is capable of performing word-by-word translation, with the difference being that the input, instead of text, is a speech segment corresponding to a spoken word in the source language.

In this paper, we propose a framework for building a ST system using only independent monolingual corpora of speech and text.
The two corpora can be collected independently which greatly reduces human labeling efforts.
Our framework starts by initializing a ST system with a cross-modal bilingual dictionary inferred from the monolingual corpora to perform word-by-word translation.
To further improve the quality of the translations, we incorporate a pre-trained language model (LM) and sequence denoising autoencoder~(DAE)~\cite{sutskever2014sequence,vincent2008extracting} that contain prior knowledge about the target language; their primary function is to consider context in lexical choices and handle local reordering and multi-aligned words.
To the best of our knowledge, this is the first work that tackles ST in an unsupervised setting.
More importantly, experiments show that our unsupervised system achieves comparable results to supervised end-to-end models~\cite{berard2018end} despite the lack of supervision.

% Before this section, we didn't mention any concepts of embedding spaces.
\section{Proposed Framework}
Our framework builds on several recently developed techniques for unsupervised speech processing and MT.
We first derive a ST system that can perform simple word-by-word translation.
Next, we integrate a language model into the framework to introduce contextual information during the translation process.
Finally, we post-process the translated results using a DAE to handle local reordering and multi-aligned words.
Below we describe each step in detail.

\subsection{Word-by-Word Translation System}
In our framework, a speech corpus from the source language is first pre-processed using an unsupervised speech segmentation algorithm~\cite{kamper2017embedded} to generate speech segments corresponding to spoken words.
We then apply a neural architecture called Speech2Vec~\cite{chung2018speech2vec,chung2017learning} to learn a speech embedding space from the set of speech segments such that each vector corresponds to a word whose semantics has been captured.
A text embedding space that captures word semantics can be learned by training Word2Vec~\cite{mikolov2013distributed} on a text corpus from the target language.
Based on the assumption that monolingual word embedding spaces are approximately isomorphic, since languages are used to convey thematically similar information in similar contexts~\cite{barone2016towards}, it is theoretically possible to align these two spaces.

To achieve this, one can use an unsupervised bilingual dictionary induction~(BDI) algorithm to learn a cross-lingual mapping from the source embedding space to the target embedding space.
Two of the most representative BDI algorithms are~$\mathrm{MUSE}$~\cite{conneau2018word} and~$\mathrm{VecMap}$~\cite{artetxe2018robust}, neither of which rely on cross-lingual signals.
Note that both these BDI algorithms were originally proposed for aligning two embedding spaces learned from text.
In~\cite{chung2018unsupervised}, however, the authors showed that~$\mathrm{MUSE}$ can also be applied to learn a \textit{cross-modal} alignment between embedding spaces learned from speech and text.
%\footnote{This is reasonable as both speech and text are expressing languages.}.
In our experiments, we include the results of both algorithms for comparison.

We obtain a rudimentary ST system after deriving a cross-modal and cross-lingual mapping from speech to the text corpora, which is essentially a linear transformation~$W$.
Given an unseen speech utterance, we first segment it into several speech segments using the speech segmentation algorithm previously mentioned.
Then, for each speech segment that potentially corresponds to a spoken word, we map it from the speech embedding space to the text embedding space via~$W$ and apply nearest neighbor search to decide its text translation.
However, the translations generated by this preliminary system are far from acceptable since nearest neighbor search does not consider the context of the current word.
In many cases, the correct translation is not the nearest target word but synonyms or other close words with morphological variations, prompting us to incorporate further improvements.

\subsection{Language Model for Context-Aware Beam Search}
We incorporate contextual information into word-by-word translation by introducing a LM during the decoding process~\cite{kim2018improving}.
Let~$w_{s}$ be the word vector mapped from speech to the text embedding space and~$w_{t}$ the word vector of a possible target word.
Given a history~$h$ of target words before~$w_{t}$, the score of~$w_{t}$ being the translation of~$w_{s}$ is computed as:
\[
LM(w_{t}; w_{s}, h) = \log \frac{f(w_{s}, w_{t}) + 1}{2} + \lambda_{\mathrm{LM}} \log p(w_{t}|h),
\]
where~$\lambda_{\mathrm{LM}}$ is the weight parameter that decides how \textit{context-aware} the system is, and~$f(w_{s}, w_{t})\in [-1, 1]$ is the cosine similarity between~$w_{s}$ and~$w_{t}$, linearly scaled to the range~$[0, 1]$ to make it comparable with the output probability of the LM.
Empirically, we found that setting~$\lambda_{\mathrm{LM}}$ to~0.1 yields the best performance.
Accumulating the scores per position, we perform a beam search to allow only reasonable translation hypotheses.

\subsection{Sequence Denoising Autoencoder}
We may achieve semantic correctness through learning an appropriate cross-modal bilingual dictionary and using a LM.
However, to further improve the quality of the translations, it is also necessary to consider syntactic correctness.
To this end, we apply a sequence DAE to correct the translated outputs.
By injecting noise to the input sequence during the training process, the DAE learns to output the original~(clean) sequence given a corrupted, noisy input.
In our framework, we adopt three noise simulation techniques proposed in~\cite{kim2018improving}: word insertion, deletion and permutation.
We seek to simulate the noise introduced during the word-by-word translation process with these three techniques.
Readers can refer to~\cite{kim2018improving} for more details.
Along with the context-aware LM, we found that adopting a DAE further boosts translation performance.
% add examples of the three techniques?

\begin{comment}
The entire framework, including the training and test flows, is summarized in Algorithm~\ref{alg:unsupervised-ST}.

\begin{algorithm}
  \caption{Unsupervised ST}
  \label{alg:unsupervised-ST}
  \begin{algorithmic}[1]
    \Procedure{Training}{speech corpus in source language, text corpus in target language}
      \State Learn speech segments correspond to spoken words
      \State Learn speech embedding space $\mathrm{S}_\mathrm{corpus}$ using Speech2Vec or $\mathrm{A}_\mathrm{corpus}$ using Audio2Vec
      \State Learn text embedding space $\mathrm{T}_\mathrm{corpus}$ using Word2Vec
      \State Learn unsupervised cross-modal bilingual dictionary $\mathcal{D}$
      \State For each word, compute $LM$
      \State Learn the sequence denoising autoencoder
      \EndProcedure
    \end{algorithmic}\\
  \midrule
  \begin{algorithmic}[1]
    \Procedure{Testing}{sentence of speech}
      \State Obtain speech segment, and then speech embedding
      \State Obtain a sequence of the corresponding text through $\mathcal{D}$
      \State For each word, compute $LM$
      \State Given a corrupted sequence of LM output, output a clean sentence
    \end{algorithmic}
  \end{algorithm}
\end{comment}

\section{Experiments}
% comparing methods
\subsection{Datasets}
We used an English-to-French speech translation dataset~\cite{kocabiyikoglu2018augmenting} augmented from the LibriSpeech ASR corpus~\cite{panayotov2015librispeech}.
The dataset is split into train, dev, and test sets; all come with a collection of English speech utterances and their corresponding French text translations.
The train set contains~100 hours of speech, which was used to train Speech2Vec~\cite{chung2018speech2vec} to obtain the speech embedding space.
For the text embedding space, we trained Word2Vec~\cite{mikolov2013distributed} on two different corpora---the parallel corpus that contains the text translations, and an independent corpus crawled from French Wikipedia.
For evaluation, we merged the dev and test sets, resulting in speech data of about~6 hours.
BLEU scores~\cite{papineni2002bleu} were used as the evaluation metric.

\subsection{Model Architectures and Training Details}
We trained Speech2Vec following the same procedure used in~\cite{chung2018unsupervised}.
The text embedding space was trained by Word2Vec using fastText~\cite{bojanowski2017enriching} with default settings without subword information.
The dimension of both speech and text embeddings is~100.
For both~$\mathrm{VecMap}$~\cite{artetxe2018robust} and~$\mathrm{MUSE}$~\cite{conneau2018word}, we followed the default settings of the implementations released by their original authors.
For the LM, we trained a 5-gram count-based LM using KenLM~\cite{heafield2011kenlm} with its default settings.
Finally, we implemented the DAE, structured as a 6-layer Transformer~\cite{vaswani2017attention}, with embedding and hidden layer size of~512, a feedforward sublayer size of~2,048, and~8 attention heads.

\subsection{Results and Discussions}
We first study the similarities between different pairs of embedding spaces to be aligned in Section~\ref{sec:eigen}.
We then present the main ST results in Section~\ref{sec:main}.
\newcommand{\myparagraph}[1]{\vspace{.4em} \noindent \textbf{#1}\ }

\subsubsection{Eigenvector Similarity}
\label{sec:eigen}
Having approximately isomorphic embedding spaces is important for BDI.
To quantify whether the embedding spaces are isomorphic, or similar in structure, we computed the eigenvector similarity, which is derived from Laplacian eigenvalues.
Both our study and~\cite{sogaard2018limitations} demonstrate that the eigenvector similarity metric is correlated to the performance of the translation task, which implies that the metric reflects the distance between embedding spaces in a meaningful way.
The similarity is computed as follows.
Let~$L_1$ and~$L_2$ be the Laplacians of two nearest neighbor embedding graphs.
We search for the smallest value of~$k$ for each graph such that the sum of largest~$k$ Laplacian eigenvalues is smaller than 90\% of the Laplacian eigenvalues.
Then, we select the smallest~$k$ across two graphs and compute the squared differences between the largest~$k$ Laplacian eigenvalues in two graphs.
The differences is the eigenvector similarity we use to measure the similarity between embedding spaces.
Note that a \textit{higher} value of the eigenvector similarity metric indicates that the given two embedding spaces are \textit{less} similar.

\begin{table}[htbp]
  \centering
  \caption{Embedding similarity of different speech and text embeddings pair evaluated by eigenvector similarity. We denote the embedding training method and corpus name in upper and lower case, respectively. For the pair, we denote the speech and text embedding space at the left and right side, respectively. For example, $\mathrm{A}_\mathrm{libri}$ - $\mathrm{T}_\mathrm{wiki}$ represents the speech embedding space trained on the LibriSpeech corpus using Audio2Vec and the text embedding space trained on Wikipedia corpus. $\mathrm{A, S, T}$ indicates Audio2Vec, Speech2Vec and text (Word2Vec) embedding.}
  \label{tab:eigen-sim}
  \resizebox{\columnwidth}{!}{
    \begin{tabular}{cc}
      \toprule
      Speech \& text embedding spaces pair  &  Eigenvector similarity   \\
      \midrule
      $\mathrm{A}_\mathrm{libri}$ - $\mathrm{T}_\mathrm{libri}$  &  14.74  \\
      $\mathrm{A}_\mathrm{libri}$ - $\mathrm{T}_\mathrm{wiki}$   &  15.02  \\
      $\mathrm{S}_\mathrm{libri}$ - $\mathrm{T}_\mathrm{libri}$  &   6.43  \\
      $\mathrm{S}_\mathrm{libri}$ - $\mathrm{T}_\mathrm{wiki}$   &   7.17  \\
      \bottomrule
    \end{tabular}
  }
\end{table}

Table~\ref{tab:eigen-sim} presents the eigenvector similarity of different speech-text pairs.
The eigenvector similarity of speech and text embedding space pairs is smaller when we trained the speech embedding using the Speech2Vec algorithm than the Audio2Vec~\cite{chung2016audio} algorithm.
These results are expected since Speech2Vec utilizes semantic context of the speech corpus, similarly to how Word2Vec uses that of the text corpus.
Furthermore, we applied skip-gram as a training methodology for both algorithms, resulting in isomorphic embedding spaces.
In contrast, Audio2Vec focuses on similarities in acoustics rather than semantics, thus the learned embedding space differs fundamentally.
Embedding space pairs learned from comparable corpora also yield higher similarity, since the word distributions are more similar; for example, the distribution of English LibriSpeech speech embeddings is more similar to that of the French LibriSpeech text embeddings than French Wikipedia text embeddings.

\subsubsection{Speech-to-text Translation}
\label{sec:main}

We present the results of our unsupervised approach as well as supervised baselines in Table~\ref{tab:main-results}.
We trained every system~10 times and report both the best and average performance.
In configurations (a-d), we replicate state-of-the-art supervised algorithms and arrived at the conclusion that cascaded systems perform better than their end-to-end counterparts and beam search performs better than greedy search.
Note that cascaded systems require more supervision than end-to-end systems, whereas our approach makes no assumptions of having speech-text or language pairs of the comparable corpora.

\begin{table}[htbp]
  \centering
  \caption{Different configurations for speech-to-text translation and their performance. The numbers in the section of unsupervised methods denoted as BLEU score (\%) of $\mathrm{VecMap}$ / BLEU score (\%) of $\mathrm{MUSE}$. The notation used in the Table is the same as Table~\ref{tab:eigen-sim}. For cascaded systems, we followed the ASR and MT pipeline in~\cite{berard2018end}. E2E stands for end-to-end.}
  \label{tab:main-results}
  \resizebox{\columnwidth}{!}{
    \begin{tabular}{cccc}
      \toprule
      \multicolumn{2}{c}{System} &  Best  &  Average  \\
      \midrule
      \midrule
      \multicolumn{4}{c}{\textit{\makecell{Cascaded and end-to-end ST systems~(supervised)}}}  \\
      \midrule
      (a) &  Cascaded + greedy   &  13.7  &  13.0  \\
      (b) &  Cascaded + beam     &  14.2  &  13.2  \\
      (c) &  E2E + greedy        &  12.3  &  11.6  \\
      (d) &  E2E + beam          &  12.7  &  12.1  \\
      \midrule
      \midrule
      \multicolumn{4}{c}{\textit{\makecell{Our alignment-based ST systems~(unsupervised)}}}  \\
      \midrule
      (e) & $\mathrm{A}_\mathrm{libri}$ - $\mathrm{T}_\mathrm{libri}$  &  0.0 / 0.0  &  0.0 / 0.0  \\
      (f) & $\mathrm{A}_\mathrm{libri}$ - $\mathrm{T}_\mathrm{wiki}$   &  0.0 / 0.0  &  0.0 / 0.0  \\
      (g) & $\mathrm{S}_\mathrm{libri}$ - $\mathrm{T}_\mathrm{libri}$  &  4.5 / 4.6  &  4.2 / 2.7  \\
      (h) & $\mathrm{S}_\mathrm{libri}$ - $\mathrm{T}_\mathrm{wiki}$   &  3.7 / 2.1  &  3.0 / 0.9  \\
      (i) & (g) + $\mathrm{LM}_{\mathrm{libri}}$   &   5.2 /  5.0  &   4.7 /  2.9  \\
      (j) & (g) + $\mathrm{LM}_{\mathrm{wiki}}$    &   9.5 /  8.8  &   9.0 /  5.7  \\
      (k) & (g) + $\mathrm{LM}_{\mathrm{wiki}}$ + $\mathrm{DAE}_{\mathrm{wiki}}$   &  12.2 / 11.8  &  11.3 /  7.3  \\
      (l) & (h) + $\mathrm{LM}_{\mathrm{wiki}}$ + $\mathrm{DAE}_{\mathrm{wiki}}$   &  11.5 / 9.1  &  10.8 /  6.2  \\
      \bottomrule
    \end{tabular}
  }
\end{table}

In configurations (e-l), we showcase the performance of our unsupervised approach, denoted as (BLEU score of $\mathrm{VecMap}$ / BLEU score of $\mathrm{MUSE}$) in the columns of Table~\ref{tab:main-results}.

\myparagraph{Alignment Quality} 
Configurations (e-h) demonstrate that eigenvector similarity of speech and text embedding space pairs have strong positive correlation, namely comparing the relative performances to those shown in Table~\ref{tab:eigen-sim}, with the BLEU score of alignment-based ST tasks.
The results, from configurations (g) and (h), illustrates that using comparable corpora, and thus better alignment, affects the quality of ST.
It also hints that there may exist a threshold of usefulness in alignment performances.
Since configurations (e) and (f) lie underneath that threshold, they achieve scores of zero.
These findings indicate that eigenvector similarity of embedding spaces could serve as an indicator of unsupervised ST performance.

\myparagraph{Unsupervised BDI} 
In all of our unsupervised experiments, we compared the performance between two unsupervised BDI algorithms, $\mathrm{VecMap}$ and  $\mathrm{MUSE}$.
$\mathrm{VecMap}$ outperforms $\mathrm{MUSE}$ in all but one experiment, demonstrating that $\mathrm{VecMap}$ can be applied to more difficult scenarios through weak, fully unsupervised initialization with iterative mapping improvements, whereas $\mathrm{MUSE}$, which maps embeddings to the shared space through adversarial training, could only succeed on a more limited set of conditions.
Additionally, $\mathrm{VecMap}$ trains more stably and faster than $\mathrm{MUSE}$, which has a similar best performance but much lower average performance.

\myparagraph{Language Model Integration} 
Integrating a LM improves the performance of ST in all experimental configurations, regardless of the selection of corpus, configurations (g) versus (i) and (j); configurations (h) versus (l) generalize this result to different embedding spaces.
By comparing configurations (i) and (j), we discover that the text corpus used to train the LM does not need to be the same as the one used for Word2Vec text embedding space training.
In fact, adopting the LM trained on the Wikipedia corpus ($\mathrm{LM}_{\mathrm{wiki}}$) produces better performance than using that trained on the LibriSpeech corpus ($\mathrm{LM}_{\mathrm{libri}}$).
Since introducing the LM grounds words into a context based on the previous word, the much larger $\mathrm{LM}_{\mathrm{wiki}}$, containing more words, topic contexts, and sentence structures, serves as a better approximation of the French language than $\mathrm{LM}_{\mathrm{libri}}$.

\myparagraph{Sequence DAE} 
In configurations (j) versus (k), we show that applying DAE on top of the baseline alignment architecture and LM can further enhance performance in unsupervised ST; the performance is now comparable to end-to-end supervised systems.
This also justifies our alignment and post-processing approach since configuration (k) essentially has the same degree of supervision as configurations (c) and (d) and performs similarly well while employing a completely different approach.
We attribute this to the DAE's ability to reconstruct corrupted data after translation.
Since the semantic alignment method we used may retrieve synonyms based on context, rather than the exact syntactically correct word ~\cite{chung2018unsupervised}, it is possible that the output even when taking the LM into account is still syntactically incorrect.
Moreover, one of the key obstacles in training Speech2Vec lies in the limited performance of unsupervised speech segmentation methods.
By incorporating a DAE, we could limit these negative effects after translation.
Last but not least, the DAE was trained on $\mathrm{LM}_{\mathrm{wiki}}$ rather than $\mathrm{LM}_{\mathrm{libri}}$.
This design decision follows from the observation of the LM corpus choice: since the DAE should learn the French language, a larger, more diverse dataset would perform better than the same dataset used for Word2Vec text embeddings.

\myparagraph{Scenario of Real-world ST}
In configuration (l), we conducted experiments modeling a real-world setting where there exists no comparable speech and text corpora.
Instead, we need to collect them independently from different sources.
Text data exists in more abundance than speech data and thus we usually adopt the text embedding learned from larger corpus such as Wikipedia, which configuration (h) replicates to our best efforts.
By comparing configurations (k) and (l), we demonstrate that the performance of our proposed framework under no supervision is only slightly inferior to the best performance achieved using unsupervised alignment, which requires  comparable corpora for speech and text embedding spaces and should be considered supervised.
The proposed unsupervised ST framework is thus promising for low language resource ST.

\section{Conclusions}
In this paper, we propose a framework capable of performing speech-to-text translation in a completely unsupervised manner.
Since the system translates using an inferred cross-modal bilingual dictionary trained without parallel data between speech and text, it could be applied to low or zero-resource languages.
By incorporating knowledge of the target language, through adding a LM and a DAE, our system greatly enhances the translation performance: We achieved comparable performance with state-of-the-art end-to-end systems using parallel corpora and only slightly lower scores without it.
These results indicate that our approach could serve as a promising first step towards fully unsupervised speech-to-text translation.
Future works include testing the proposed framework on other language pairs and examining the relationship between embedding quality and translation performance in more detail.
% (real) completely unsupervised ST: what if we use out-of-domain speech corpus for training?

\vfill\pagebreak

\bibliographystyle{IEEEbib}
\bibliography{strings,refs}

\begin{thebibliography}{10}

\bibitem{weiss2017sequence}
Ron Weiss, Jan Chorowski, Navdeep Jaitly, Yonghui Wu, and Zhifeng Chen,
\newblock ``Sequence-to-sequence models can directly translate foreign
  speech,''
\newblock in {\em INTERSPEECH}, 2017.

\bibitem{bansal2018low}
Sameer Bansal, Herman Kamper, Karen Livescu, Adam Lopez, and Sharon Goldwater,
\newblock ``Low-resource speech-to-text translation,''
\newblock in {\em INTERSPEECH}, 2018.

\bibitem{berard2018end}
Alexandre B{\'e}rard, Laurent Besacier, Ali~Can Kocabiyikoglu, and Olivier
  Pietquin,
\newblock ``End-to-end automatic speech translation of audiobooks,''
\newblock in {\em ICASSP}, 2018.

\bibitem{berard2016listen}
Alexandre B{\'e}rard, Olivier Pietquin, Laurent Besacier, and Christophe
  Servan,
\newblock ``Listen and translate: A proof of concept for end-to-end
  speech-to-text translation,''
\newblock in {\em NIPS Workshop on End-to-end Learning for Speech and Audio
  Processing}, 2016.

\bibitem{lample2018phrase}
Guillaume Lample, Myle Ott, Alexis Conneau, Ludovic Denoyer, and Marc'Aurelio
  Ranzato,
\newblock ``Phrase-based \& neural unsupervised machine translation,''
\newblock in {\em EMNLP}, 2018.

\bibitem{artetxe2018unsupervised}
Mikel Artetxe, Gorka Labaka, and Eneko Agirre,
\newblock ``Unsupervised statistical machine translation,''
\newblock in {\em EMNLP}, 2018.

\bibitem{conneau2018word}
Alexis Conneau, Guillaume Lample, Marc'Aurelio Ranzato, Ludovic Denoyer, and
  Herv{\'e} J{\'e}gou,
\newblock ``Word translation without parallel data,''
\newblock in {\em ICLR}, 2018.

\bibitem{artetxe2018robust}
Mikel Artetxe, Gorka Labaka, and Eneko Agirre,
\newblock ``A robust self-learning method for fully unsupervised cross-lingual
  mappings of word embeddings,''
\newblock in {\em ACL}, 2018.

\bibitem{sennrich2016improving}
Rico Sennrich, Barry Haddow, and Alexandra Birch,
\newblock ``Improving neural machine translation models with monolingual
  data,''
\newblock in {\em ACL}, 2016.

\bibitem{chung2018unsupervised}
Yu-An Chung, Wei-Hung Weng, Schrasing Tong, and James Glass,
\newblock ``Unsupervised cross-modal alignment of speech and text embedding
  spaces,''
\newblock in {\em NIPS}, 2018.

\bibitem{sutskever2014sequence}
Ilya Sutskever, Oriol Vinyals, and Quoc Le,
\newblock ``Sequence to sequence learning with neural networks,''
\newblock in {\em NIPS}, 2014.

\bibitem{vincent2008extracting}
Pascal Vincent, Hugo Larochelle, Yoshua Bengio, and Pierre-Antoine Manzagol,
\newblock ``Extracting and composing robust features with denoising
  autoencoders,''
\newblock in {\em ICML}, 2008.

\bibitem{kamper2017embedded}
Herman Kamper, Karen Livescu, and Sharon Goldwater,
\newblock ``An embedded segmental k-means model for unsupervised segmentation
  and clustering of speech,''
\newblock in {\em ASRU}, 2017.

\bibitem{chung2018speech2vec}
Yu-An Chung and James Glass,
\newblock ``Speech2vec: A sequence-to-sequence framework for learning word
  embeddings from speech,''
\newblock in {\em INTERSPEECH}, 2018.

\bibitem{chung2017learning}
Yu-An Chung and James Glass,
\newblock ``Learning word embeddings from speech,''
\newblock in {\em NIPS Workshop on Machine Learning for Audio Signal
  Processing}, 2017.

\bibitem{mikolov2013distributed}
Tomas Mikolov, Ilya Sutskever, Kai Chen, Greg Corrado, and Jeff Dean,
\newblock ``Distributed representations of words and phrases and their
  compositionality,''
\newblock in {\em NIPS}, 2013.

\bibitem{barone2016towards}
Antonio Valerio~Miceli Barone,
\newblock ``Towards cross-lingual distributed representations without parallel
  text trained with adversarial autoencoders,''
\newblock in {\em RepL4NLP}, 2016.

\bibitem{kim2018improving}
Yunsu Kim, Jiahui Gend, and Hermann Ney,
\newblock ``Improving unsupervised word-by-word translation with language model
  and denoising autoencoder,''
\newblock in {\em EMNLP}, 2018.

\bibitem{kocabiyikoglu2018augmenting}
Ali Kocabiyikoglu, Laurent Besacier, and Olivier Kraif,
\newblock ``Augmenting librispeech with french translations: A multimodal
  corpus for direct speech translation evaluation,''
\newblock in {\em LREC}, 2018.

\bibitem{panayotov2015librispeech}
Vassil Panayotov, Guoguo Chen, Daniel Povey, and Sanjeev Khudanpur,
\newblock ``Librispeech: an {ASR} corpus based on public domain audio books,''
\newblock in {\em ICASSP}, 2015.

\bibitem{papineni2002bleu}
Kishore Papineni, Salim Roukos, Todd Ward, and Wei-Jing Zhu,
\newblock ``Bleu: A method for automatic evaluation of machine translation,''
\newblock in {\em ACL}, 2002.

\bibitem{bojanowski2017enriching}
Piotr Bojanowski, Edouard Grave, Armand Joulin, and Tomas Mikolov,
\newblock ``Enriching word vectors with subword information,''
\newblock {\em Transactions of the Association for Computational Linguistics},
  vol. 5, pp. 135--146, 2017.

\bibitem{heafield2011kenlm}
Kenneth Heafield,
\newblock ``Kenlm: Faster and smaller language model queries,''
\newblock in {\em WMT}, 2011.

\bibitem{vaswani2017attention}
Ashish Vaswani, Noam Shazeer, Niki Parmar, Jakob Uszkoreit, Llion Jones, Aidan
  Gomez, {\L}ukasz Kaiser, and Illia Polosukhin,
\newblock ``Attention is all you need,''
\newblock in {\em NIPS}, 2017.

\bibitem{sogaard2018limitations}
Anders S{\o}gaard, Sebastian Ruder, and Ivan Vuli{\'c},
\newblock ``On the limitations of unsupervised bilingual dictionary
  induction,''
\newblock in {\em ACL}, 2018.

\bibitem{chung2016audio}
Yu-An Chung, Chao-Chung Wu, Chia-Hao Shen, Hung-Yi Lee, and Lin-Shan Lee,
\newblock ``Audio word2vec: Unsupervised learning of audio segment
  representations using sequence-to-sequence autoencoder,''
\newblock in {\em INTERSPEECH}, 2016.

\end{thebibliography}

\end{document}